\newif\ifsingle
\documentclass[a4paper,journal]{IEEEtran}
\IEEEoverridecommandlockouts

\usepackage[top=1.9cm, bottom=4.3cm,left=1.3cm,right=1.3cm]{geometry}
\def\BibTeX{{\rm B\kern-.05em{\sc i\kern-.025em b}\kern-.08em T\kern-.1667em\lower.7ex\hbox{E}\kern-.125emX}}
\usepackage[noadjust]{cite}
\usepackage[T1]{fontenc}
\usepackage[latin9]{inputenc}
\usepackage{color}
\usepackage{amsmath}
\usepackage{amssymb}
\usepackage{graphicx}
\usepackage{subcaption}
\usepackage[normalem]{ulem} 
\usepackage{textcomp}
\usepackage{xcolor}
\usepackage{balance}
\usepackage{setspace}
\usepackage{lipsum}
\usepackage{multirow}
\usepackage{algorithm}
\usepackage{algpseudocode}
\usepackage[bookmarks,hidelinks]{hyperref}
\allowdisplaybreaks
\usepackage{bbm}
\usepackage{pgfplots}
\usepackage{soul}
\usepackage{ulem}

\usepackage[compact]{titlesec}
\titlespacing{\section}{0pt}{2ex}{1ex}
\titlespacing{\subsection}{0pt}{1ex}{0ex}
\titlespacing{\subsubsection}{0pt}{0.5ex}{0ex}

\makeatletter
\ifsingle
 
\else
 
\fi 
\definecolor{NewColor}{rgb}{0.2,0,0.5}

\begin{document}
\title{Seeing Through WiFi: Lightweight Human Pose Estimation with Dynamic Kernel Attention}

 \author{
	\IEEEauthorblockN{Toan D. Gian, Van-Dinh Nguyen,  Vo Phi Son, Nhan Thanh Nguyen,  Dinh Thai Hoang, \\
 Diep N. Nguyen, Nguyen Cong Luong and Symeon Chatzinotas
 \vspace{-15pt}  }
 

\thanks{T. D. Gian and V. P. Son are with the Smart Green Transformation Center (GREEN-X), VinUniversity, Vietnam. V.-D. Nguyen (corresponding author) is with Trinity College Dublin,  Ireland  (e-mail: dinh.nguyen@tcd.ie). 
N. T. Nguyen is with the University of Oulu, Finland, (email: nhan.nguyen@oulu.fi).
D. T. Hoang and D. N. Nguyen are with University of Technology Sydney,  Australia (e-mail: \{hoang.dinh, diep.nguyen\}@uts.edu.au)
N. C. Luong is with PHENIKAA
University,  Vietnam (e-mail: luong.nguyencong@phenikaa-uni.edu.vn).
 S. Chatzinotas is with  SnT, University of Luxembourg, Luxembourg (e-mail: Symeon.Chatzinotas@uni.lu).
Part of this work was presented at \textit{IEEE International Conference on Communications and Electronics},  July-August, 2024 \cite{GianICCE24}.}

}

\maketitle

\begin{abstract}
WiFi-based human pose estimation (HPE) enables the detection and interpretation of human body positions and movements without the need for wearable devices while preserving individual privacy concerns. Implementing this solution requires enhancing model performance and maintaining efficiency, especially on resource-constrained devices. This paper introduces a novel framework, WiLHPE, for lightweight and efficient human pose estimation using WiFi CSI signals. Empowered by a camera-based model during training, WiLHPE processes raw WiFi signals directly to estimate human poses in the testing phase. It employs a novel neural network architecture to dynamically learn convolutional kernels and apply attention mechanisms across channel and frequency spaces. This innovative method diversifies the kernels to improve the recognition capabilities of WiFi signals without adding complexity, ensuring efficiency. Additionally, the Tree-Structured Parzen Estimator algorithm is employed to optimize the critical hyperparameters of the neural network efficiently, minimizing the time required for optimal hyperparameter search compared to heuristic methods. Results from experiments on both the MM-Fi and WiPose datasets highlight the superiority of WiLHPE over state-of-the-art approaches, achieving $85.96\%$ and $94.27\%$ at $\text{PCK}_{50}$, respectively, with minimal computational overhead. Notably, WiLHPE performs impressively even under challenging conditions, maintaining around $80\%$ at $\text{PCK}_{50}$ under AWGN noise with an error variance of $0.5$. 
\end{abstract}


\section{Introduction}
Recent technological advancements in analyzing human behavior have led to innovations that enhance daily life. A promising area is daily human activity monitoring \cite{surveyHAR1,surveyHAR2}. Traditional methods using visual sensors or wearable devices have notable success but face challenges. Wearable sensors \cite{wearableHAR1,wearableHAR2} are often cumbersome and uncomfortable for prolonged or high-intensity use. Visual sensors \cite{videoHAR1,videoHAR2} struggle with fixed angles, occlusions, glare, and low-light conditions, compromising image quality. Additionally, privacy concerns and regulations limit the feasibility of camera-based pose estimation where privacy is essential.

Sensor-based approaches, particularly those using WiFi CSI, have emerged as promising alternatives for human pose estimation (HPE), addressing limitations of conventional camera-based methods. For instance, RF-Pose in \cite{Zhao2019ThroughWallHM} and \cite{RFskeleton} used radio frequency signals to generate skeletal representations of the human body, showing remarkable results, especially when bodies are occluded. Additionally, an infrared-based approach \cite{infrared} has proven effective for pose estimation under diverse lighting conditions, including both bright and dark environments. Despite these advancements, the adoption of these techniques faces significant hurdles due to stringent requirements, such as complex hardware (\textit{e.g.} a $16$+$4$ intricately designed and synchronized T-shaped antenna array) and demanding RF signals (\textit{e.g.} frequency modulated continuous wave with a broad signal bandwidth of $1.78$ GHz). The implementation of these approaches necessitates expensive equipment, challenging their practical application for everyday use.

Recently, WiFi-based approaches in \cite{wifiHAR1} and \cite{wifiHAR2} have emerged as promising solutions to these challenges. By leveraging ubiquitous WiFi signals, bistatic configuration methods provide a non-intrusive, privacy-aware alternative to traditional approaches. Electromagnetic interactions between WiFi signals and the human body, including reflection, refraction, and penetration, enable the extraction of pose features from channel state information (CSI) for HPE. However, real-world deployment remains challenging. CSI often lacks the granularity needed for fine-grained pose estimation, mixing human motion with environmental effects and producing unrealistic postures. Moreover, maintaining continuity and smoothness in synthesized skeleton movements is essential for reliable activity representation and usability.

\subsection{Motivation}
The widespread utilization of WiFi infrastructure and open-source software has significantly increased interest in exploiting CSI for human sensing \cite{open-source,HARscenes}. Many works have particularly focused on enhancing the accuracy of human sensing through WiFi-based HPE. To facilitate deployment and cost-effectiveness, the authors in \cite{Toward2020} leveraged readily available WiFi devices to perform effective HPE tasks. Guo \textit{et al.} \cite{Guo2020} demonstrated that off-the-shelf WiFi devices could capture 2D images of human skeletons. However, this approach is limited to single-perspective poses, resulting in suboptimal performance when annotations are constrained.
In contrast, the work in \cite{Toward2020} introduced WiPose, the first 3D HPE system relying on the CSI amplitude, and has shown superior performance in controlled experimental settings. However, WiPose requires subjects to remain stationary during experiments and depends on nine distributed antennas, restricting its practicality for everyday applications such as smart homes and health monitoring. Meanwhile, the authors in \cite{Wang2019CanWE} used two standard WiFi devices to collect CSI data, utilizing 30 subcarriers, and integrated these devices with a fully convolutional network to predict 2D human poses. Nevertheless, the key findings indicate that limited subcarrier resolution compromises result precision, highlighting the need for a WiFi acquisition system with more antennas and subcarriers. This improvement, though, introduces increased computational complexity when processing high-dimensional input with deep neural networks.

Recently, a novel approach called MetaFi was introduced in \cite{MetaFi} to address the low resolution of subcarriers by utilizing commodity WiFi devices equipped with $114$ subcarriers. Although this approach achieved impressive resolution, its effectiveness in HPE tasks was limited by constraints within its neural network architecture, particularly learnability and complexity. Recently, the MetaFi++ solution in \cite{MetaFi++2023} employed commodity WiFi devices featuring one transmission and three receiver antennas, each supporting $114$ subcarriers. The associated neural network imposed a substantial computational burden with approximately $26.42$ million parameters, compromising its scalability for large-scale implementation. Consequently, there remains a critical need to develop a high-resolution, easy-to-deploy commodity WiFi system, coupled with a high-performance yet lightweight neural network capable of efficiently processing high-dimensional WiFi data. These advancements are essential to unlocking the full potential of WiFi-based human sensing in real-world applications.
\begin{figure}[t]
  \centering
  \includegraphics[width=0.75\linewidth]{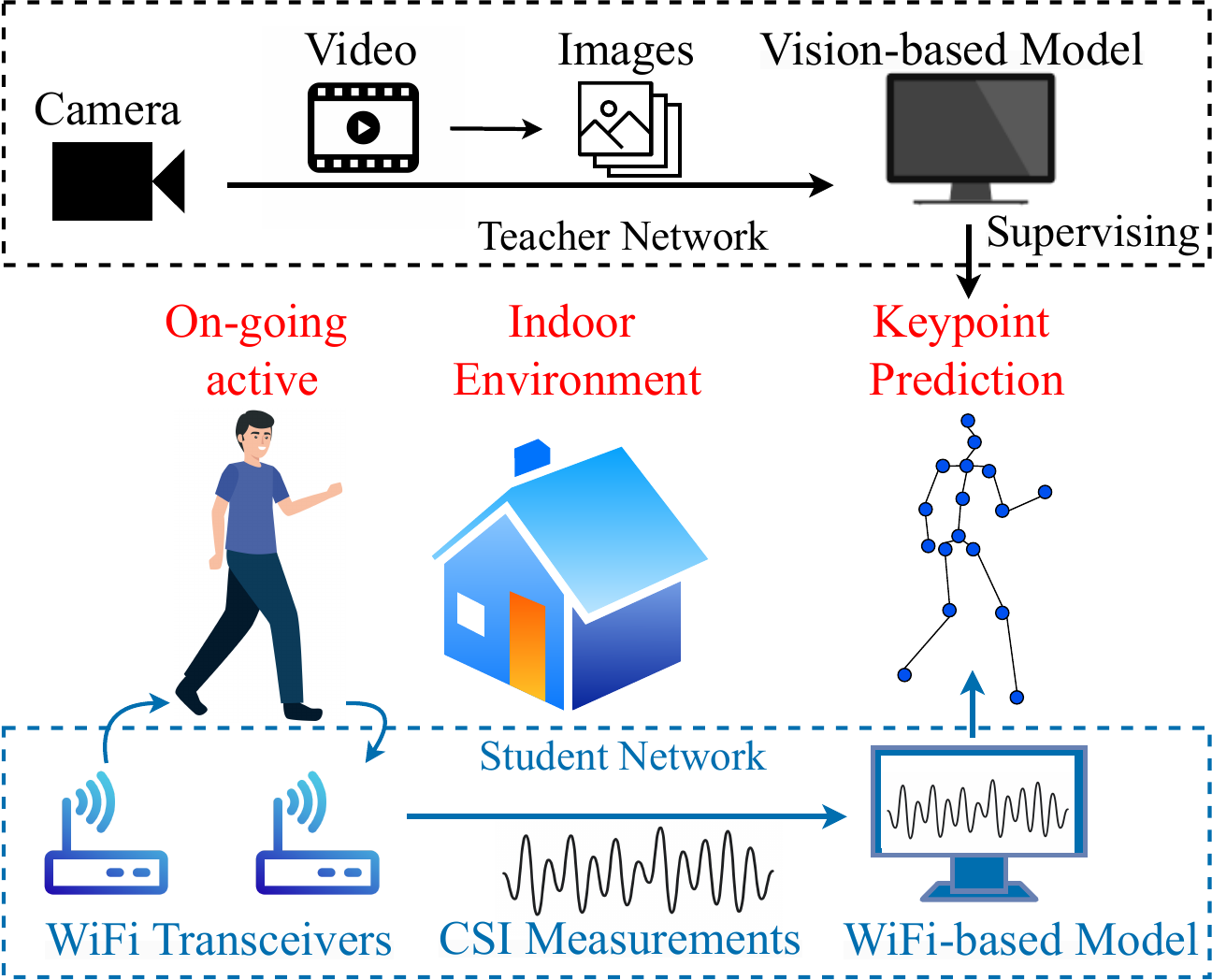}
  \caption{Illustration of indoor WiFi-based human sensing: $i$) the teacher network extracts labels from RGB images, and $ii$) the student network learns from CSI data to predict human keypoints in real time.}
  \label{fig:figure1}
\end{figure}

\subsection{Main Contributions}
To overcome the aforementioned shortcomings, we propose a novel WiFi-based network called ``WiLHPE,'' designed to generate the \underline{l}ightweight and efficiently \underline{h}uman \underline{p}ose \underline{e}stimation from \underline{Wi}Fi CSI signals and RGB images. As shown in Fig. \ref{fig:figure1}, WiLHPE adopts a teacher-student architecture with two networks. A pre-trained vision-based teacher extracts pose labels from RGB images to supervise training, while the WiFi-based student predicts poses from raw WiFi signals. During testing, the student functions as a real-time pose estimator without requiring cameras.

 The student network employs a channel-frequency dynamic convolution network (CF-DyNet) to learn representations from CSI efficiently. Within CF-DyNet, we introduce CF-DyConv, a convolutional architecture with multi-dimensional attention that weights input features by importance to focus on informative patterns. In CSI systems, data consists of frequency amplitude sequences across multiple antennas and short time frames (\textit{e.g.}, $10$ ms in MM-Fi \cite{yang2023mmfi}). Although temporal resolution is limited, the channel domain provides spatial diversity to resolve occlusions, while the frequency domain captures fine-grained spectral signatures related to joint positions. Therefore, our architecture jointly exploits channel and frequency information, adaptively optimizing convolutional kernels for improved performance.

Our main contributions can be summarized as follows:
\begin{itemize}
     \item Within the WiLHPE system, we introduce CF-DyNet sub-networks to effectively process WiFi CSI signals for human pose estimation tasks. This framework includes a novel CF-DyConv architecture that utilizes dynamic convolutional filters to adjust their kernel size based on input data characteristics. The introduced neural network is lightweight, requires no pre-processing, has high learning capability, and features a simple structure.
    \item We \textcolor{black}{leverage} the Tree-Structured Parzen Estimator (TPE) \cite{TPE} method to optimize the critical hyperparameters of the considered CF-DyNet sub-networks, avoiding reliance on heuristics. This allows WiLHPE to achieve optimal results without extensive hyperparameter search time.
   \item We perform intensive experiments to demonstrate the effectiveness of the proposed WiLHPE framework, achieving $85.96\%$ and $94.27\%$ at $\text{PCK}_{50}$ on MM-Fi \cite{yang2023mmfi} and WiPose \cite{PerUnet2022} datasets, respectively, while balancing model performance and computational cost in human pose estimation tasks. Additionally, WiLHPE with the proposed CF-DyConv architecture exhibits robust performance on the MM-Fi dataset under challenging conditions, outperforming traditional CNNs. 
\end{itemize}

\subsection{Paper Organization and Mathematical Notations}
The rest of this paper is organized as follows. Section \ref{subsec:sec2} provides the literature review on HPE tasks and attention-based convolution. Section \ref{subsec:sec3} gives the background of the CSI and dynamic convolution. The proposed WiLHPE framework is presented in Section \ref{subsec:sec4}. Section \ref{subsec:sec5} details HPE datasets, evaluation metrics, results, and discussions, and Section \ref{subsec:sec6} concludes the paper.  

\textit{Mathematical notations:} Throughout this paper, matrices and vectors are written as bold uppercase and lowercase letters, respectively, while the scalar number is denoted in lowercase.  $\odot$ and $*$ denote the multiplication operation and the convolution operation, respectively. $\delta(\cdot)$ is the Dirac delta function.  The notation $x \sim \mathcal{CN}(0,\sigma^2)$ implies that $x$ is a circularly-symmetric complex Gaussian random variable with zero mean and variance  $\sigma^2$. $\|\cdot\|$ represents the vector's Euclidean norm. Finally, $\mathbb{R}$ denotes a set of real numbers.

\section{Related Work}
\label{subsec:sec2}

\noindent \textbf{WiFi sensing}: Leveraging the widespread availability of WiFi infrastructure, numerous studies have explored the capabilities of WiFi devices over recent decades. This has enabled various sensing tasks such as indoor localization and tracking \cite{tracking1,tracking2}, walking speed estimation \cite{walking1,walking2}, inference of breathing and heartbeats \cite{breathing1,breathing2}, human identification \cite{iden1,iden2}, human presence and movement detection \cite{movement1,movement2}, and gesture recognition \cite{ABLSTM,oneshot}. 
However, these tasks are generally considered coarse-grained person perception tasks, as they primarily detect broad actions or states without delving into detailed body dynamics. In contrast, our proposed WiLHPE framework focuses on finer-grained personal perception tasks, which involve more detailed and precise analyses of human activities. For instance, we aim to represent human pose landmarks, allowing us to capture specific body postures and movements, thereby advancing the capabilities of human pose estimation. We emphasize that our work focuses solely on HPE tasks, rather than other WiFi sensing tasks, due to the differing dataset characteristics associated with each task. 

\noindent \textbf{Human pose estimation}: In the field of computer vision (CV), 2D HPE has been extensively explored using powerful deep learning techniques and abundantly annotated datasets from standard cameras \cite{Cao2018OpenPoseRM, RMPE2017}. Additionally, the works  \cite{HumanEva2010} and \cite{Microsoft2012} have implemented specialized devices to enhance the estimation performance of human motion. Despite the impressive results achieved by vision-based methods, their deployment faces significant challenges, including poor lighting, occlusion, blurred images, and privacy concerns. Efforts to address privacy issues using light-based methods (\textit{e.g.} \cite{Li2015HumanSU}) have been made, but these methods exhibit limited performance in dark or occluded conditions.
While LiDAR has shown fine-grained person detection capabilities \cite{VoxNet2015}, the high cost and power requirements limit its practicality for daily and household use. In contrast, the WiFi-based approaches are capable of addressing these limitations by ensuring privacy and maintaining functionality in various lighting and occlusion scenarios. In particular, RF-based strategies have emerged as promising solutions to overcome these challenges. For instance, RFCapture \cite{adib2015} can outline human bodies even through walls, while RFPose \cite{Zhao2018} and its 3D variant \cite{RFskeleton} can extract 2D and 3D skeletons from RF signals with the aid of visual data. Nevertheless, these methods often require sophisticated hardware (\textit{e.g.} a $16$+$4$ intricately designed and synchronized T-shaped antenna array) and specific signal conditions (\textit{e.g.} frequency modulated continuous wave with a broad signal bandwidth of $1.78$ GHz), limiting their practical applications. In contrast, our approach framework demonstrates that robust HPE can be achieved using accessible, high-resolution WiFi systems and lightweight neural network architectures, making it more feasible for real-world deployment.

\noindent \textbf{Attention-based CNN}: Very recently, there has been extensive research on developing novel CNN architectures to enhance the performance of traditional CNNs, particularly through the use of attention mechanisms. A representative work is SENet \cite{SENet2017}, which introduced the Squeeze-and-Excitation (SE) module to model interdependencies among feature channels. Building on SE's aggregation-recalibration design, \cite{DynamicConv} proposed Dynamic Convolution (DyConv), dynamically combining multiple kernels based on input attention. However, DyConv relies only on channel-domain information, limiting kernel-space modeling. To address this, ODConv \cite{ODConv} and F-DyConv \cite{FDyconv} were developed as more advanced variants. Similarly, SKNet \cite{SKNets} fuses multi-scale features using attention across different kernel sizes, but still focuses mainly on domain-specific information, reducing versatility. F-SKNet further improved SKNet, particularly for speaker verification. In HPE-based wireless sensing, recent works in \cite{toan, toan1, toan2, toan3} employed a kernel-selective CNN to reduce model complexity by focusing on informative Wi-Fi signals, whereas \cite{toan4} considered this setting in an edge-server architecture.

In this work, we propose a convolutional architecture that goes beyond channel-only processing by exploiting spatial-spectral features in both channel and frequency dimensions of short-frame CSI data. Given CSI's limited temporal granularity, our attention mechanism selectively extracts discriminative features across these domains, enabling robust and efficient representation learning.

\section{Preliminaries\label{subsec:sec3}}

\subsection{Channel State Information (CSI)}
In WiFi-based networks, CSI captures the unique characteristics of a communication channel, providing detailed insights into how WiFi signals propagate in a specific physical environment. This includes various phenomena such as multipath effects, reflections, diffraction, and scattering. Modern commercial WiFi devices, following the IEEE 802.11 standard, utilize OFDM at the physical layer and often employ multiple antennas. CSI data meticulously records the amplitude attenuation and phase shift of multipath signals across each communication subcarrier for every antenna pair. This detailed information allows for the detection of subtle human movements that affect WiFi signal propagation. The frequency domain represents the signal as the channel impulse response, enabling precise analysis and optimization of wireless communication performance, which can be expressed as
\begin{equation}
h(\tau)=\sum_{l=1}^{L}\alpha_{l}e^{j\phi_l}\delta(\varphi-\varphi_{l})
\end{equation}
where $\alpha_l$, $\phi_l$, and $\varphi$ are the amplitude, phase, and time delay of the $l$-th multipath component, respectively; $L$ indicates the total number of multipath components given in the channel; and $\delta(\varphi)$ denotes the Dirac delta function.

\subsection{Overview of Dynamic Convolution}

\begin{figure}
    \centering
    \includegraphics[width=1.0\linewidth]{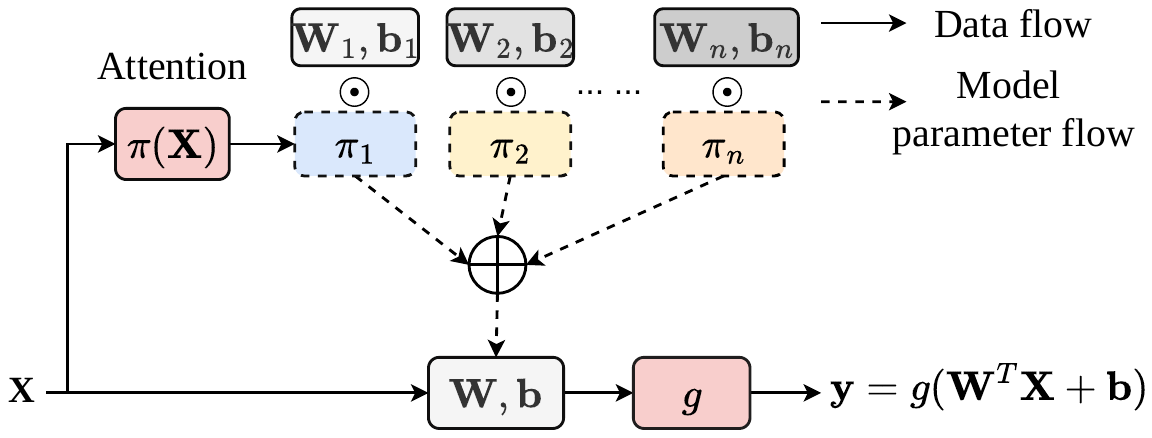}
    \caption{Overview of Dynamic Convolution.}
    \label{fig:ovr_Dyconv}
\end{figure}
\noindent \textbf{Basic Concept.} 
Dynamic convolution (DyConv) was introduced in \cite{DynamicConv} to enhance the representational capacity of conventional convolution by dynamically adjusting the convolution kernel in response to input characteristics. Firstly, DyConv extracts attention weights based on the input and then performs a weighted summation of $n$ basis kernels using these attention weights to derive the optimal kernel for the input, as shown in Fig.~\ref{fig:ovr_Dyconv}. Given the input and output channel numbers, denoted as $C_{\mathtt{in}}$ and ${C_{\mathtt{out}}}$, respectively, DyConv can be mathematically expressed as:
\begin{equation}
\textbf{Y}=g(\textbf{W}^{T}\textbf{X}+\textbf{b})=(\sum_{i=1}^{n}\alpha_{i}(\textbf{W}_{i} +\textbf{b}_i))\ast\textbf{X}\label{eq: basicDC}
\end{equation}
where $\alpha_{i}=\pi_i(\textbf{X})$,  $\textbf{X}\in\mathbb{R}^{H\times W\times C_{\mathtt{in}}}$ and $\textbf{Y}\in\mathbb{R}^{H\times W\times C_{\mathtt{out}}}$  denote the input features and output features with $H$ and $W$ being the height and width of the channel, respectively. Herein, $\mathbf{W}_i \in\mathbb{R}^{H\times W\times C_{\mathtt{in}} C_{\mathtt{out}}}$ is the $i$-th convolutional kernel, $\alpha_{i}\in \mathbb{R}$ is the attention
weight for the $i$-th convolutional kernel which is computed by the attention function $\pi_{i}(\textbf{X})$ conditioned on the input features, and $\ast$ denotes the convolution operation.
\begin{figure*}[t]
  \centering
  \includegraphics[width=.75\linewidth]{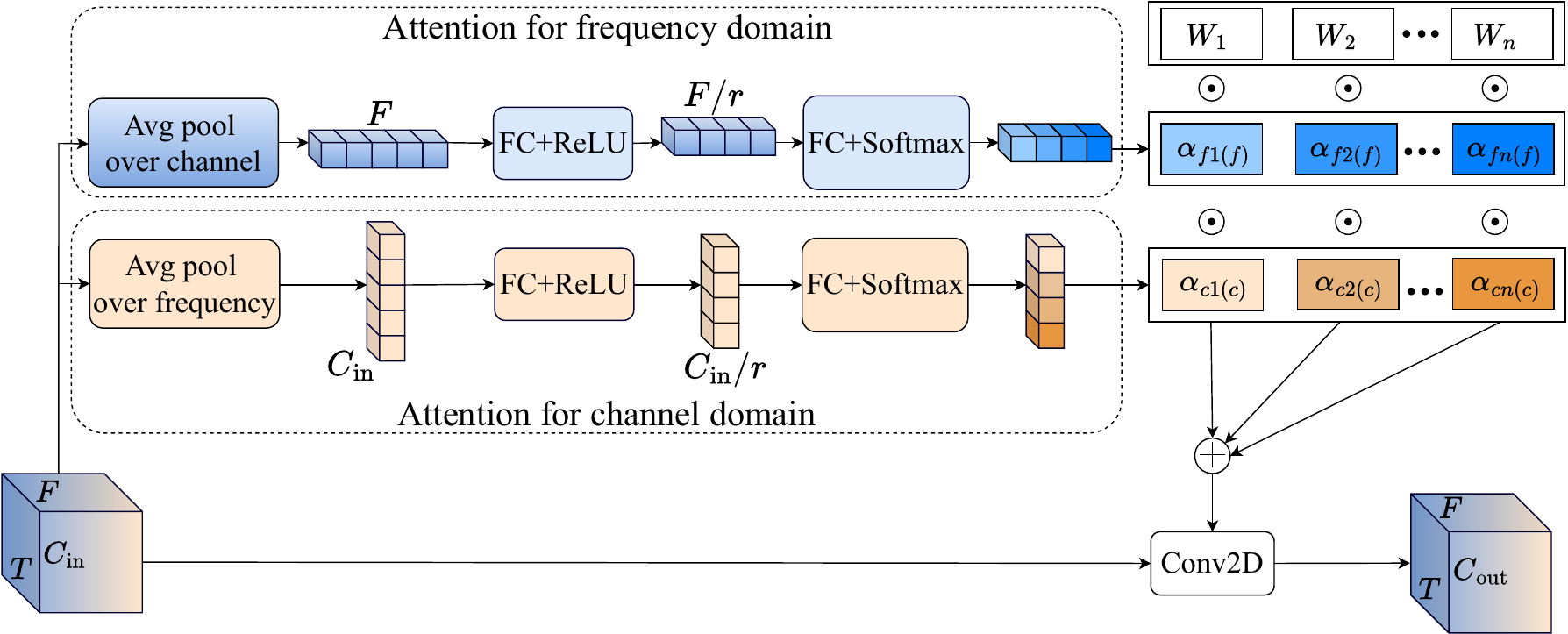}
  \caption{The proposed WiLHPE network: Empowered by the CV model, the student network predicts human pose from CSI data. The data is processed through CNN and CF-DyNet blocks, with features passed to the Decoder to generate human keypoints, as discussed in Section \ref{subsec:Network architecture}.}
  \label{fig:figure2}
\end{figure*}

\noindent \textbf{Limitation Discussions.}
According to \eqref{eq: basicDC}, DyConv comprises two key elements: Convolutional kernels $\mathbf{W}_i$, and attention function $\pi_{i}(\textbf{X})$ used to calculate their attention scalars $\alpha_{i}$. With $n$ convolutional kernels, the kernel space is characterized by four dimensions: The spatial kernel size $k \times k$, input channel number $C_{\mathtt{in}}$, output channel number $C_{\mathtt{out}}$ for each kernel, and the total number of kernels, $n$. We note that $\pi_{i}(\textbf{X})$ assigns a single attention scalar $\alpha_{i}$ to each $\mathbf{W}_i$, resulting in all $C_{\mathtt{out}}$ filters $\mathbf{W}_{i}^{m}\in\mathbb{R}^{k\times k \times C_{\mathtt{in}}}$ with $m \in \{1,\cdots, C_{\mathtt{out}}\}$ sharing the same attention value for the input $\textbf{X}$. In simpler terms, DyConv disregards the spatial, input channel, and output channel dimensions of $\mathbf{W}_i$, focusing solely on one remaining dimension. This narrow focus might result in inefficient utilization of the kernel space when devising attention mechanisms to empower $n$ dynamic convolutional kernels. Meanwhile, each frame of WiFi-CSI data exists concurrently in both the frequency and channel domains, signifying pose information (\textit{i.e.} frequency) from a specific viewpoint. The physical environment predominantly influences each frequency component independently at a channel level. Consequently, traditional DyConv may struggle to capture features from WiFi data effectively.

\section{Proposed WiLPHE Framework \label{subsec:sec4}}

We first describe CSI preprocessing, then present the CF-DyConv implementation and complexity analysis. Finally, we introduce the WiLHPE architecture, including hyperparameter optimization and learning objectives.

\subsection{Channel-Frequency Dynamic Convolution (CF-DyConv)}

CF-DyConv is a key component for building an efficient neural network that learns and weights kernels based on contextual cues using diverse attention mechanisms. As illustrated in Fig.~\ref{fig:figure2}, the CF-DyConv architecture is divided into two distinct flows. The first flow generates new weights for 2D convolution by applying attention mechanisms in the frequency and channel domains. Initially, the input data undergoes average pooling along the frequency dimension, followed by FC layers with ReLU and softmax activations to produce frequency attention weights. A similar process is applied in the channel domain to generate channel attention weights. These attention weights are then combined to create new weights for the 2D convolutional layer.
The second flow uses these newly obtained weights to perform 2D convolution on the input data, extracting relevant features efficiently. \vspace{1pt}

\noindent \textbf{Pre-Processing CSI Data.}~Let us start by considering the CSI dataset which contains $G$ samples denoted as $\{C_{i}\text{\}}_{i=1}^{G},$ where $C\in\mathbb{R}^{T_{x}\times R_{x}\times F\times T}$ with $T_x$, $R_x$ and $F$ are the numbers of transmitting antennas, receiving antennas and subcarriers, respectively, while $T$ is the frame time. To simplify, we consider the $T_x \times R_x $ dimension as the input channels ${C_{\mathtt{in}}}$.   Consequently, the CSI data is represented as  \textbf{X} $\in$ $\mathbb{R}^{F \times T \times {C_{\mathtt{in}}}}$, where $F$ and $T$ dimensions represent the values in the frequency domain and the time domain, respectively.

\noindent \textbf{CF-DyConv.}~For a given \textbf{X}, we adopt a SENet \cite{SENet2017} including the squeeze and excitation stages. SENet generates attention scalars for convolutional kernels along the channel and frequency dimensions, denoted ${\alpha_{c}(\mathbf{X})}$ and ${\alpha_{f}(\mathbf{X})}$, respectively, as shown in Fig. \ref{fig:figure2}. The squeeze stage uses the global average pooling (GAP) to obtain the global spatial information, which encapsulates the global information into the channel-wise feature vector $\textbf{z}_c \in \mathbb{R}^{C_{\mathtt{in}}}$, such as $\text{GAP}_{\texttt{\texttt{chan}}}$, and frequency-wise feature vector $\textbf{z}_f \in \mathbb{R}^F$, such as $\text{GAP}_{\texttt{\texttt{freq}}}$. The channel-wise  and frequency-wise feature vectors are computed as
\begin{align}
\textbf{z}_c&=\text{GAP}_{\texttt{chan}}(\textbf{X})=\frac{1}{F\times T}\sum_{f=1}^{F}\sum_{\text{t=1}}^{T}\textbf{X}(f,t)
\label{eq:GAP_channel}\\
\textbf{z}_f&=\text{GAP}_{\texttt{freq}}(\textbf{X})=\frac{1}{C\times T}\sum_{c=1}^{C}\sum_{\text{t=1}}^{T}\textbf{X}(c,t)
\label{eq:GAP_freq}
\end{align}
respectively. The excitation stage then leverages the information aggregated during the squeeze stage by using two FC layers with activation functions. These layers generate normalized attention weights for $n$ convolution kernels, acting as a gating mechanism to capture the nonlinear interactions among multiple channels. The process can be summarized as follows:
\begin{align}
\alpha_c &= f_{\texttt{excitation}}(\mathbf{z}_{c}) =\sigma\Big(\mathbf{W}_{c2}\delta\left(\mathbf{W}_{c1} \mathbf{z}_c/\tau\right)\Big)
 \label{eq:post-process2}\\
\alpha_f &= f_{\texttt{excitation}}(\mathbf{z}_{f}) =\sigma\Big(\mathbf{W}_{f2}\delta\left(\mathbf{W}_{f1} \mathbf{z}_f/\tau\right)\Big)
\label{eq:post-process3}
\end{align}
where $\sigma$ and $\delta$ are the softmax and ReLU activation functions, respectively; $\mathbf{W}_{c1} \in \mathbb{R}^{\frac{C_{\mathtt{in}}}{r} \times C_{\mathtt{in}}}$, $\mathbf{W}_{c2} \in \mathbb{R}^{C_{\mathtt{in}} \times \frac{C_{\mathtt{in}}}{r}}$,  $\mathbf{W}_{f1} \in \mathbb{R}^{\frac{F}{r} \times F}$ and $\mathbf{W}_{f2} \in \mathbb{R}^{F \times \frac{F}{r}}$ are the learnable weights of the excitation stage. The reduction ratio $r$ acts as a bottleneck to
reduce the input dimensionality of the first FC
layer. The temperature $\tau$ is utilized to smooth the attention distribution.  The attention weights $\alpha_c$ and $\alpha_f$ are fed to the input $\textbf{X}$ to produce more informative kernels. Note that $\alpha_c$ is an attention scalar that weights filters according to their importance in capturing channel information, while $\alpha_f$ weights filters based on their significance in capturing frequency information. By leveraging this attention mechanism,  we enhance the kernel's ability to capture features that are most informative and best suited for the specific dataset.

To generate convolutional kernels that capture diverse information, we combine channel and frequency scalars with the kernels after learning the attention and then aggregate the kernels. As shown in Fig. \ref{fig:figure2}, the attention weights ${\alpha_c}$ and  $\alpha_f$ are multiplied with convolution kernels, providing diversity to the kernels. 
The output feature $\mathbf{Y}\in\mathbb{R}^{F \times T \times C_{\mathtt{out}}}$ is computed as
\begin{equation}
\mathbf{Y}=\mathbf{X}\ast\Big(\sum_{i=1}^{n}\alpha_{ci}\odot\alpha_{fi}\odot\mathbf{W}_{i}\Big)\label{eq:post-process4}
\end{equation}
where $n$ is the number of convolution kernels, $\mathbf{W}_{i}$ is the $i$-th convolution kernels, and $\odot$ denotes the multiplication operations.

\noindent \textbf{Complexity Analysis.}~The proposed CF-DyConv introduces two additional computational stages: $i)$ computing the attention weights
$\alpha_c(\mathbf{X})$ and $\alpha_f(\mathbf{X})$, and $ii)$ aggregating stages. In particular, the computation cost of the first stage is estimated as \( FT C_{\mathtt{in}} + 2 C_{\mathtt{in}}^2/r + 2 F^2/r + n C_{\mathtt{in}} /r +  n F /r  \)  Mult-Adds, while the second stage aggregates $n$ convolution kernels with kernel size $D_k \times D_k$,
$C_{\mathtt{in}}$ input channels and $C_{\mathtt{out}}$ output channels requires \( n C_{\mathtt{in}} C_{\mathtt{out}} D_k^2 + n C_{\mathtt{out}} \)  Mult-Adds with \( D_k \) being the kernel size. Therefore, the total computational cost of the proposed CF-DyConv is \( FT C_{\mathtt{in}} + 2 C_{\mathtt{in}}^2/r + 2 F^2/r + n C_{\mathtt{in}} /r + n F /r  \) $+$ \( n C_{\mathtt{in}} C_{\mathtt{out}} D_k^2 + n C_{\mathtt{out}} \) Mult-Adds. In comparison, basic convolution requires $FTC_{\mathtt{in}}C_{\mathtt{out}}D_k^2$  Mult-Adds. Thus, the additional cost of our approach is negligible if $n\ll FT$. 

We note that the two key benefits of CF-DyConv to enhance the overall network performance of the WiLHPE framework are:
\begin{itemize}
\item$\noindent\textbf{Comprehensive Exploitation}$: CF-DyConv captures both frequency and channel information by applying dynamic convolutional filters to each frame of the input. This comprehensive approach enables the model to extract a wide range of features from the data, significantly enhancing performance in HPE tasks.
\item$\noindent\textbf{Dynamic Convolutional Filters}$: CF-DyConv utilizes an attention-based mechanism to adaptively adjust its weights based on the input data. This dynamic adjustment allows the model to learn more effective feature representations with minimal additional computational cost.
\end{itemize}

\begin{figure*}[t]
  \centering
  \includegraphics[width=0.75\linewidth]{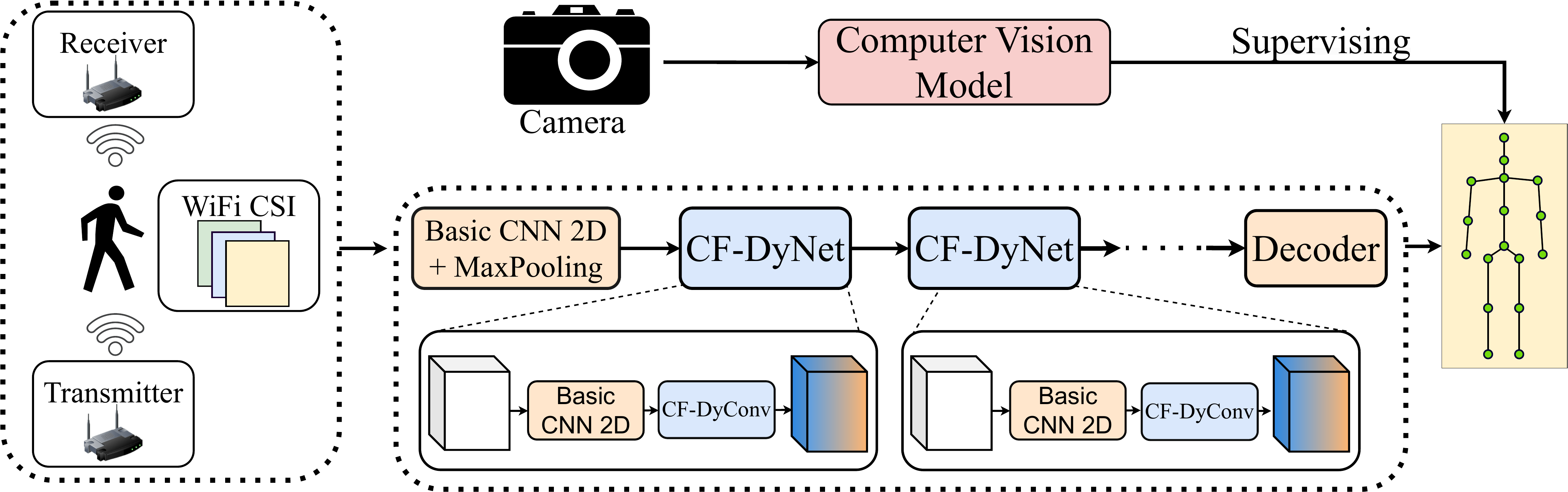}
  \caption{Block diagram of the proposed channel-frequency dynamic convolution.}
  \label{fig:figure3}
\end{figure*}

\subsection{WiLHPE-based Network Architecture \label{subsec:Network architecture}}
\begin{figure}
    \centering
    \includegraphics[width=1.0\linewidth]{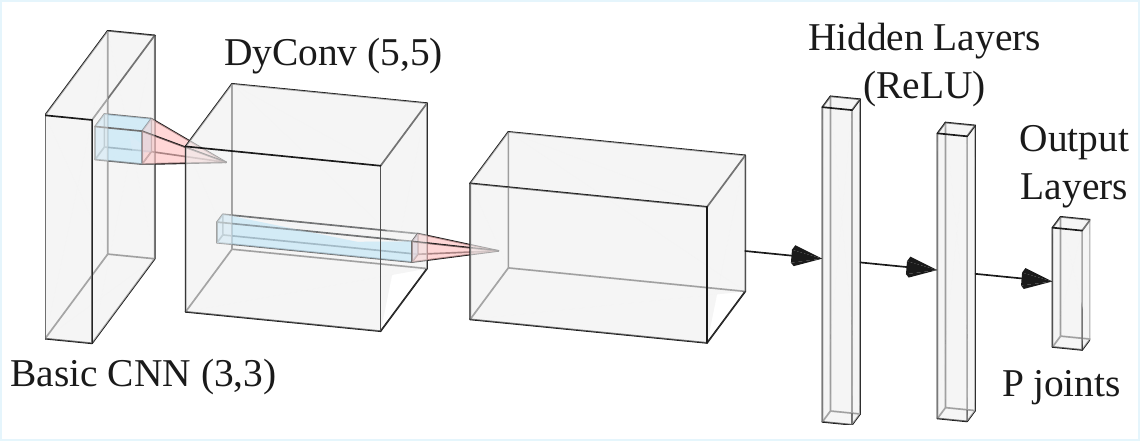}
    \caption{The architecture of CF-DyNet and decoder blocks (assuming $M=1$).}
    \label{fig:figure4}
\end{figure}

As shown in Fig. \ref{fig:figure3}, the teacher network employs a computer vision model, such as HRNet-w48 \cite{yang2023mmfi,MetaFi++2023} to extract 2D key points from multi-view frames. These 2D key points serve as the ground truth $\mathbf{y}$ for estimating human poses across various modalities, including pixel coordinates $(a, b)$ represented as $\textbf{y}=\{(a_{i},b_{i})|i\in[1,\cdots,P]\}$, where $P$ denotes the number of key points. The student network is a neural network designed to predict human poses under the supervision of the teacher network. Specifically, the student network is trained using supervised learning, with the true labels provided by the teacher network's ground truth. In Fig. \ref{fig:figure4}, this network primarily comprises basic CNNs, $M$  repeated sub-networks known as CF-DyNet to extract the feature transformation network. In particular, CF-DyNet consists of a basic CNN and a CF-DyConv, each with distinct input channels denoted as $n_1$ and $n_2$, respectively. Each CNN block is followed by an AvgPooling layer of size 2. The features extracted from CF-DyNet are then passed to the decoder module, which includes an FC layer with a ReLU activation function and an output size of $2P$, as illustrated in Fig. \ref{fig:figure4}. This block generates predictions of the human body's key points $\hat{\mathbf{y}}=\{(\hat{a}_{i},\hat{b}_{i})|i\in[1,\cdots, P]\}$. \vspace{-5pt}

\noindent \textbf{TPE Algorithm}: As previously mentioned, two key parameters influence performance: The reduction ratio $r$ and the temperature $\tau$. The former adjusts the capacity and computational cost of the attention module, while the latter in \eqref{eq:post-process2} and \eqref{eq:post-process3} is used to control the sparsity of the weights. The selection of these parameters and their effects are explored in the ablation experiments. In addition, to avoid manually tuning the hyperparameters of WiLHPE, we customize the TPE algorithm \cite{TPE} to automatically search the hyperparameters. Instead of evaluating $p(y|x)$ as in the traditional algorithms, TPE models $p(x|y)$ and $p(y)$. By two such densities, TPE defines $p(x|y)$ as follows:
\begin{align}
p(x|y) =
\begin{cases} 
\ell(x) & \text{if } y < y^*, \\
g(x) & \text{if } y \geq y^*.
\end{cases}
\end{align}
The TPE algorithm identifies $y^*$ as a specific quantile $\gamma$ of the observed $y$ value, ensuring that $p(y < y^*) = \gamma$, where $p(y)$ is the probability distribution of $y$. The algorithm aims to classify sample points into the superiority set $g(x)$ and the inferiority set $\ell(x)$. The optimal parameters are determined through iterative successive updates to these sets, with the goal of maximizing the Expected Improvement (EI) function, which is defined as
\begin{align}
\text{EI}_{y^*}(x) &= \frac{\gamma y^* \ell(x) - \ell(x) \int_{-\infty}^{y^*} p(y) \, dy}{\gamma \ell(x) + (1 - \gamma)g(x)}\nonumber\\
&\propto \left( \gamma + \frac{g(x)}{\ell(x)}(1-\gamma) \right)^{-1}\label{eq:post-process11}
\end{align} 
following the Bayes transformation rule \cite{TPE}. Equation \eqref{eq:post-process11} implies that maximizing EI requires minimizing the ratio $\frac{g(x)}{\ell(x)}$. Thus, each iteration returns a set of values for $x$ that decrease  $g(x)$ and increase $\ell(x)$. The hyperparameters in the set are evaluated based on the objective function. This process is then iteratively repeated to complete the hyperparameter search. The customized TPE algorithm is summarized in Algorithm \ref{alg:TPE}.

\begin{algorithm}[t]
\caption{The Tree-Structured Parzen Estimator Algorithm}
\label{alg:TPE}
\textbf{Input:} 
The search target $T$ (hyperparameters of the neural network, \textit{i.e.} number of CNN layers, batch size, and convolution kernel size), search scope $S$ (\textit{i.e.} given the integer values), and maximum number of iterations $Q$.

\textbf{Output:} Results for each search target $T$.

\begin{enumerate}
  \item Minimize the score of the objective function for $T$ and the output. 
  
  \item Obtain a couple of observations (scores) using a randomly selected set of $S$.
  
  \item Rely on scores to sort the obtained observations, and then divide them into groups based on some quantile $\gamma$. 
  
  \item Compute two densities $\ell(x_1)$ and $g(x_2)$ using Parzen Estimators.
  
  \item Select sample hyperparameters from $\ell(x_1)$, assess them using $\ell(x_1) g(x_2)$, and identify the set that produces the lowest value under $\ell(x_1) g(x_1)$ with the highest EI. These hyperparameters are then evaluated by the objective function.
  \item Update the observation list from Step 3.
  \item Repeat Steps 3-6 with a fixed number of trials or until reaching the time limit.
\end{enumerate}
\end{algorithm}

\subsection{Learning Objective\label{subsec:Learning-objective}}
We employ the mean-squared error (MSE) between the prediction and the ground truth produced by the visual pose estimation model as the loss function, defined as follows:
\begin{equation}
\mathcal{L}_{\mathtt{MSE}} = \lVert \hat{\mathbf{y}} - \mathbf{y} \rVert_2^2
 \label{eq:post-process5}
\end{equation}
where $\mathbf{\hat{y}}$ and $\mathbf{y}$ are the predicted value and target value, respectively. Instead of using the pose adjacency matrix as done in the previous studies \cite{Wang2019CanWE}, we found that the MSE loss function yields better performance. This choice is motivated by the high granularity of the CSI data. Both the MM-Fi and WiPose datasets contain numerous subcarriers of CSI data for each antenna pair, resulting in higher-resolution data. Consequently, the task becomes more manageable even with a simple loss function.

\begin{table*}[h]
  \caption{Detail of MM-Fi and WiPose Datasets.  
  }
  \label{tab:table1}
  \centering
  \normalsize
\begin{tabular}{ccccccc}
\hline 
\textbf{Dataset}\, & \textbf{Activity$\times$Subject}\, & \textbf{Tx$\times$Rx$\times$Subcarrier$\times$Time} \, & \textbf{Frequency} \, & \textbf{Packets} \, & \textbf{Training}  \,& \textbf{Testing}\tabularnewline
\hline 
\textbf{MM-Fi} & $27\times40$ & $1\times3\times314\times10$ & $5$ GHz & $320$K & $70\%$ & $30\%$\tabularnewline
\hline 
\textbf{WiPose} & $12\times12$ & $3\times3\times30\times5$ & $5$ GHz & $166$K & $70\%$ & $30\%$\tabularnewline
\hline 
\end{tabular}
\end{table*}

\section{Results and Discussions\label{subsec:sec5}}
To evaluate WiLHPE's performance, we first describe the experimental setup, including the dataset, evaluation metrics and implementation details. We then conduct a comparative analysis between WiLHPE and the current state-of-the-art (SOTA) approaches to assess its effectiveness. Finally, we conduct the robustness analysis and ablation study to demonstrate the impact of CF-DyConv.


\subsection{Experimental Setup}
\label{5a}
\subsubsection{Dataset}

We use two HPE datasets: MM-Fi \cite{yang2023mmfi} and WiPose \cite{PerUnet2022}. A brief description of both datasets is provided in Table~\ref{tab:table1}.
\begin{itemize}
\item \textbf{MM-Fi:} encompasses $17$ skeleton points of pose annotations, 
derived from both camera sensor and WiFi CSI data.
This dataset involves $40$ human subjects participating in $27$ action categories, collecting from $14$ daily activities and $13$ rehabilitation exercises.
Furthermore, MM-Fi implements three protocols: $i)$ Protocol 1 (P1) focuses on $14$ daily activities, such as picking up objects and raising arms, $ii)$ Protocol 2 (P2) involves $13$ activities performed in fixed locations, like limb extensions and $iii)$ Protocol 3 (P3) includes all $27$ activities mentioned above. Each protocol utilizes two data splitting strategies: $i)$ Setting 1 (S1) randomly divides all video samples into training and testing sets with a ratio of $3:1$ and $ii)$ Setting 2 (S2) splits the data by subject, allocating $32$ subjects for training and $8$ for testing.
    
 \item \textbf{WiPose:} contains $166$k packets with pose annotations for $18$ skeleton points. 
 CSI data is captured from $12$ actions (\textit{i.e.} wave, walk, throw, run, push, pull, jump, crouch, circle, sit down, stand up and bend) of $12$ participants.
 The data collection system is equipped with a $5$ GHz wireless router and a WiFi receiving host, acquiring CSI data with $30$ subcarriers per antenna pair, resulting in each video frame corresponding to five CSI data. The training and testing sets are randomly chosen from all data.
\end{itemize} 

\subsubsection{Evaluation Metrics}
We evaluate the system performance using three widely used metrics in HPE tasks. 
\begin{itemize} 
\item Percentage of Correct Keypoint (PCK) \cite{PCK} measures the proportion of correctly detected keypoints that fall within a certain distance threshold  of the ground truth keypoints $\alpha$, which is expressed as
\begin{equation}
    \text{PCK}_{\alpha} = \frac{1}{KP}\sum_{i=1}^{K}\sum_{j=1}^{P}I\left ( \frac{\left \| pd_{i,j} - gt_{i,j} \right \|_{2}^{2} }{ \sqrt{rs^2 + lh^2} } \leq \alpha  \right )
\label{eq:post-process6}
\end{equation}
where $I(\cdot)$ represents a binary indicator function, $K$ is the number of test frames, and $j \in \{1, 2, \ldots, P\}$ denotes the index for body joints. The distance $\lVert pd_{i,j} - gt_{i,j} \rVert^2_2$ signifies the Euclidean distance between the predicted results ($pd$) and the ground truth ($gt$) at the $j$-th body joint in the $i$-th test frame. Next, the term $\sqrt{rs^2 + lh^2}$,  known as the torso length, measures the distance between the right shoulder ($rs$) and the left hip ($lh$) and is used to normalize the prediction error. Given the indicator function $I(\cdot)$ and the stricter threshold $\alpha$, the higher the PCK score, the better the network performance.

\item Mean Per Joint Position Error (MPJPE) \cite{PCK}, measured in millimeters (mm), evaluates the average error between the predicted and ground-truth positions for all joints after aligning the pelvis of the estimated and true pose. Therefore, a lower MPJPE metric indicates better network performance. The MPJPE is calculated as
\begin{equation}
\text{MPJPE} = \frac{1}{KP} \sum_{i=1}^{K} \sum_{j=1}^{P} \lVert pd_{i,j} - gt_{i,j} \rVert_2
. \label{eq:post-process1}
\end{equation}

\item Procrustes-Aligned Mean Per Joint Position Error (PA-MPJPE), which is actually the MPJPE after aligning the predicted results to the ground truth through a Procrustes transformation, provides an additional alignment method to improve the accuracy of 3D body joint position estimation compared to only using basic MPJPE.
\end{itemize} 

\subsubsection{Benchmark Schemes}
For performance comparison, we consider several SOTA WiFi-based approaches. In particular, we consider MetaFi++ \cite{MetaFi++2023}, which is currently recognized as the most effective model for the HPE task on MM-Fi. Other well-established HPE methods are implemented, including Wi-Pose \cite{Toward2020}, Wi-Mose \cite{Wang2021}, WiSPPN \cite{Wang2019CanWE} and WiLDAR \cite{WiLDAR2024}.

\subsubsection{Adding Noise} To analyze the impact of unfavorable conditions on WiLHPE, we introduce additional noise into the testing data to assess our model's performance. Specifically, we employ Additive White Gaussian Noise (AWGN), denoted as $e$, where $e \sim \mathcal{CN}(0,\sigma_{e}^{2})$ with $\sigma_{e}^{2}$ being the variance to simulate natural randomness typical in real-world scenarios.

We further investigate the impact of noise through adversarial attacks. Specifically, we implement the Fast Gradient Sign Method (FGSM) \cite{FGSM}, which manipulates input images to deceive the model by introducing carefully crafted noise. Here, we apply a single-step, untargeted perturbation (FGSM-U). This method aims to increase the network loss for a given input-label pair $(\textbf{I}, \textbf{y})$ by adding noise with a bounded $l_\infty$ norm. The noise is computed as the scaled  $\varepsilon$ sign of the gradient with respect to the objective, resulting in a perturbed image $\mathbf{I}^p$:
\begin{equation}
\mathbf{I}^p = \mathbf{I} + \varepsilon \cdot \text{sign}(\nabla_{\mathbf{I}} \mathcal{L}(f(\mathbf{I}, \theta), \mathbf{y}))
\label{eq:post-process12}
\end{equation}
and iterative un-targeted perturbation FGSM-U-N. This process involves $N$ iterations to produce the final perturbed image $\textbf{I}^p$ starting from \textbf{I}. The perturbed image $\textbf{I}^p_i$ at the $i$-th iteration for un-targeted is given as:
\begin{equation}
\mathbf{I}_{i}^{p} = C\varepsilon\Big(\mathbf{I},\, \mathbf{I}_{i-1}^p + \alpha.\text{sign}\big(\nabla_{\mathbf{I}_{i-1}^p} \mathcal{L}(f(\mathbf{I}^{p}_{i-1}, \theta)\mathbf{y})\big)\Big)
\label{eq:post-process13}
\end{equation}
where $C_{\varepsilon}(x_0, x)$ clips $x$ to $[x_0 - \varepsilon, x_0 + \varepsilon]$. The iterative attacks are generally more damaging compared to single-step attacks because they can distort the image in a highly non-linear manner over multiple iterations.

\subsubsection{Implementation Details} 
All schemes are implemented using PyTorch, leveraging an Intel 13-core i9-13900K CPU (3 GHz) and a GeForce RTX 4070 GPU. The dataset is divided into three parts: 70\% for training, 15\% for validation, and $15\%$ for testing. The model is trained for 50 epochs using either the stochastic gradient descent (SGD) with momentum algorithm or the Adam optimizer. The batch size, learning rate, and momentum are set to $32$, $0.001$ and $0.9$, respectively. These parameters are optimized using the TPE algorithm, as outlined in Algorithm~\ref{alg:TPE}. Additionally, the learning rate follows a lambda decaying schedule, gradually decreasing from the initial value to 0.

For the student network, we start by implementing a standard CNN with a kernel size of $3$ and a MaxPooling block with a size of $2$. Next, we incorporate three identical CF-DyNet blocks ($M$ = $3$), as illustrated in Fig. \ref{fig:figure4}. Each CF-DyNet block consists of a basic CNN with a kernel size of $3$ and number channels $n_1$ of $64$, followed by a CF-DyConv layer with a kernel size of $5$, $n$ (as defined in Eq. \eqref{eq:post-process4}) being set to $3$, and number channels $n_2$ of $128$. Within each CF-DyNet block, an AvgPooling block with a size of $2$ is used to reduce the feature size while preserving feature quality. The decoder consists of FC layers and ReLU activation functions, as shown in Fig. \ref{fig:figure4}, with hidden layers containing $64$ units.

\subsection{Performance Evaluation}
\label{subsec:berPer} 
\begin{table*}[!htb]
    \caption{Performance of WiLHPE on MM-Fi and WiPose Datasets. `L.' Denotes Left And
`R.' Denotes Right.\label{tab:table3}}
    \begin{subtable}{.5\linewidth}
      \centering
        \caption{MM-Fi Dataset}
        \label{tab:table3a}
        \begin{tabular}{c|cccc}
\hline 
\textbf{Keypoint} & $\textbf{PCK}_{20}$ & $\textbf{PCK}_{30}$ & $\textbf{PCK}_{40}$ & $\textbf{PCK}_{50}$ $\uparrow$\tabularnewline
\hline 
Bottom Torso & 72.03 & 84.25 & 91.61 & 94.08\tabularnewline
L.Hip & 70.08 & 83.04 & 90.84 & 94.61\tabularnewline
L.Knee & 69.98 & 83.81 & 90.87 & 94.62\tabularnewline
L.Foot & 68.19 & 83.25 & 90.38 & 94.51\tabularnewline
R.Hip & 72.79 & 85.88 & 91.36 & 94.96\tabularnewline
R.Knee & 70.86 & 83.84 & 90.33 & 94.81\tabularnewline
R.Foot & 67.95 & 80.71 & 88.72 & 92.56\tabularnewline
Center Torso & 68.45 & 82.56 & 89.41 & 93.23\tabularnewline
Upper Torso & 61.52 & 78.86 & 86.21 & 90.09\tabularnewline
Neck Base & 53.41 & 72.25 & 82.61 & 88.44\tabularnewline
Center Head & 53.02 & 72.86 & 85.59 & 88.57\tabularnewline
R.Shouder & 61.58 & 77.75 & 86.33 & 91.42\tabularnewline
R.Elbow & 33.67 & 54.41 & 68.56 & 76.44\tabularnewline
R.Hand & 8.08 & 19.34 & 30.59 & 46.83\tabularnewline
L.Shoulder & 60.27 & 77.42 & 86.13 & 91.37\tabularnewline
L.Elbow & 32.23 & 53.37 & 69.38 & 77.97\tabularnewline
L.hand & 7.03 & 17.12 & 31.21 & 46.97\tabularnewline
 & &  &  & \tabularnewline
\hline 
\textbf{Average} & \textbf{54.88} & \textbf{70.36} & \textbf{79.22} & \textbf{85.96}\tabularnewline
\hline 
\end{tabular}
    \end{subtable}%
    \begin{subtable}{.5\linewidth}
      \centering
        \caption{WiPose Dataset}
        \label{tab:table3b}
          \begin{tabular}{l|cccc}
\hline 
\textbf{Keypoint}  & $\textbf{PCK}_{5}$& $\textbf{PCK}_{10}$ & $\textbf{PCK}_{20}$ & $\textbf{PCK}_{50}$$\uparrow$\tabularnewline
\hline 
Nose & 93.75 & 96.87 & 96.87 & 96.87\tabularnewline
Neck & 90.62 & 93.75 & 96.87 & 96.87\tabularnewline
R.Shoulder & 90.62 & 96.87 & 96.87 & 96.87\tabularnewline
R.Elbow & 68.75 & 90.62 & 96.87 & 96.87\tabularnewline
R.Wrist & 71.87 & 84.37 & 96.87 & 96.87\tabularnewline
L.Shoulder & 81.25 & 90.62 & 96.87 & 96.87\tabularnewline
L.Elbow & 46.87 & 71.87 & 81.25 & 96.87\tabularnewline
L.Wrist & 56.25 & 71.87 & 78.12 & 90.62\tabularnewline
R.Hip & 68.75 & 93.75 & 96.87 & 100.00\tabularnewline
R.Knee & 71.87 & 93.75 & 100.00 & 100.00\tabularnewline
R.ankle & 75.00 & 84.37 & 96.87 & 96.87\tabularnewline
L.Hip & 50.00 & 78.12 & 87.50 & 100.00\tabularnewline
L.Knee & 59.37 & 75.00 & 93.75 & 100.00\tabularnewline
L.ankle & 71.87 & 84.37 & 93.75 & 93.75\tabularnewline
R.Eye & 96.87 & 96.87 & 96.87 & 96.87\tabularnewline
L.Eye & 43.75 & 59.37 & 78.12 & 93.75\tabularnewline
R.Ear & 93.75 & 96.87 & 96.87 & 96.87\tabularnewline
L.Ear & 25.00 & 25.00 & 28.12 & 50.00\tabularnewline
\hline 
\textbf{Average} & \textbf{69.79} & \textbf{82.46} & \textbf{89.41} & \textbf{94.27}\tabularnewline
\hline 
\end{tabular}
    \end{subtable} 
\end{table*}

\begin{table}[t]
  \caption{The MPJPE and PA-MPJPE Results of WiLHPE with Different Protocols and Settings on MM-Fi (Best in \textbf{bold} and second best in \underline{underlined})
  }
  \label{tab:table4}
  \centering
  \normalsize
\begin{tabular}{c|ccccc}
\hline 
\multirow{2}{*}{$\textbf{P}$} & \multicolumn{5}{c}{$\textbf{Setting 1}$}\tabularnewline
\cline{2-6} \cline{3-6} \cline{4-6} \cline{5-6} \cline{6-6} 
 & Packet & Training & Testing & MPJPE$\downarrow$ & PA-MPJPE$\downarrow$\tabularnewline
\hline 
$\textbf{1}$ & 166K & 116K & 50K & $\underline{142.77}$ & 93.89\tabularnewline
$\textbf{2}$ & 154K & 108K & 46K & 157.42 & $\underline{88.18}$\tabularnewline
$\textbf{3}$ & 320K & 224K & 96K & $\mathbf{140.71}$ & $\mathbf{92.39}$\tabularnewline
\hline 
\multirow{2}{*}{$\textbf{P}$} & \multicolumn{5}{c}{$\textbf{Setting 2}$}\tabularnewline
\cline{2-6} \cline{3-6} \cline{4-6} \cline{5-6} \cline{6-6} 
 & Packet & Training & Testing & MPJPE$\downarrow$ & PA-MPJPE$\downarrow$\tabularnewline
\hline 
$\textbf{1}$ & 166K & 133K & 33K & $\underline{182.16}$ & 93.22\tabularnewline
$\textbf{2}$ & 154K & 123K & 31K & 187.98 & $\mathbf{86.18}$\tabularnewline
$\textbf{3}$ & 320K & 256K & 64K & $\mathbf{179.24}$ & $\underline{93.16}$\tabularnewline
\hline 
\end{tabular}
\end{table}

\begin{figure*}[t]
  \centering
  \includegraphics[width=0.8\linewidth]{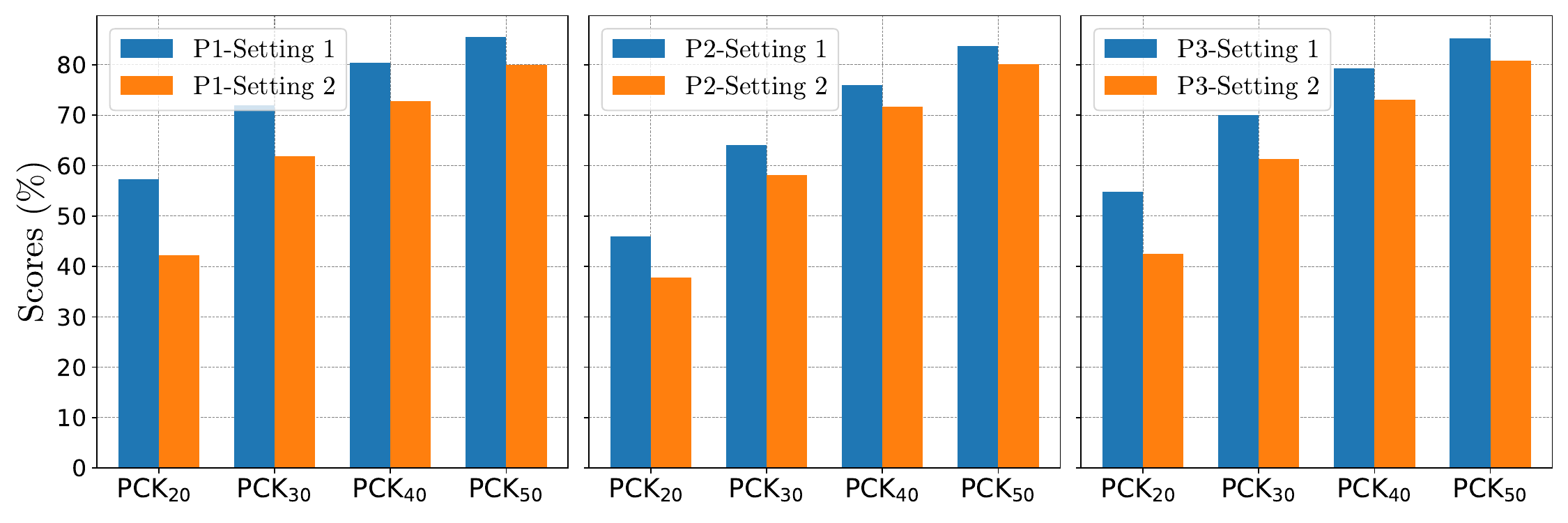}
  \caption{The $\text{PCK}_{\alpha}$ performance with different protocols and settings on MM-Fi.}
  \label{fig:figure5}
\end{figure*}

\begin{figure*}[tb]
  \centering
  \includegraphics[width=0.8\linewidth]{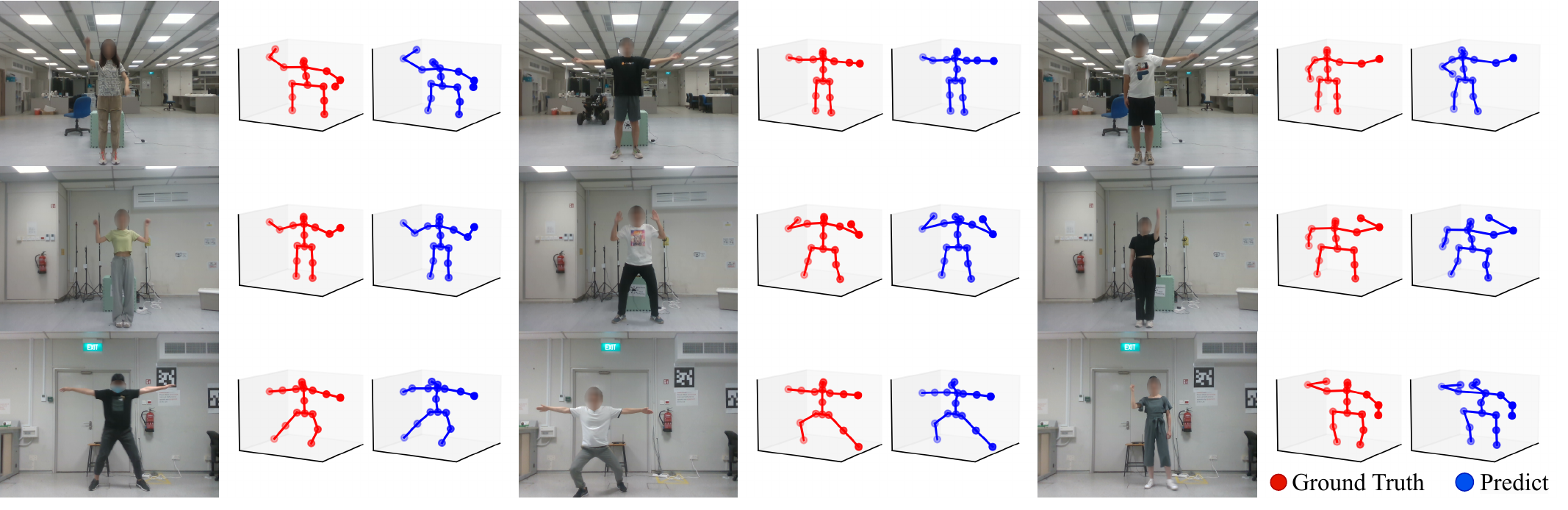}
  \caption{Visualization of the human pose landmarks generated by the vision model (red)
and WiFi model (blue) on the MM-Fi dataset.}
  \label{fig:figure6}
\end{figure*}

\subsubsection{Results on MM-Fi} We first evaluate the performance of WiLHPE on the MM-Fi dataset using the $\text{PCK}_{20}$, $\text{PCK}_{30}$, $\text{PCK}_{40}$, and $\text{PCK}_{50}$ metrics. Note that we omit the results for lower PCK thresholds due to the limitations in annotation quality and benchmark consistency of the MM-Fi dataset, as discussed in \cite{yang2023mmfi}. As shown in Table~\ref{tab:table3a}, WiLHPE exhibits impressive overall pose estimation performance, with average scores of $79.22\%$ and $85.96\%$ for $\text{PCK}_{40}$ and $\text{PCK}_{50}$, respectively. Specifically, our model shows exceptional precision in estimating the lower part of the human body (\textit{e.g.} torso, hips, knees, and feet), achieving over 94\% accuracy for $\text{PCK}_{50}$. Notably, the accuracy of these specific body joints remains about $70\%$ under more stringent conditions at $\text{PCK}_{20}$. These results are influenced by the activities performed in MM-Fi, which primarily focus on upper body movements. Consequently, challenges arise in predicting parts of the body, such as hands and elbows, where movements occur with high intensity.

In Fig. \ref{fig:figure5} and Table \ref{tab:table4}, we further evaluate the effectiveness of WiLHPE using $\text{PCK}_{\alpha}$, MPJPE, and PA-MPJPE metrics across different protocols. As depicted in Fig. \ref{fig:figure5}, WiLHPE performs better performance in \textbf{P3-S1} with $\text{PCK}_{\alpha}$, while showing the least effectiveness in \textbf{P2-S2}. As observed in Table \ref{tab:table4}, WiLHPE achieves optimal results in \textbf{P3-S1} with an MPJPE of $140.71$ mm, and the weakest performance in \textbf{P2-S2} at approximately $187$ mm. However, our model achieves favorable PA-MPJPE results in \textbf{P2-S2}, with a value of 86.18mm. These results underscore the considerable influence of protocol configurations on the WiLHPE's performance.

\begin{table*}[t]
  \caption{Performance Comparison Between WiLHPE and Benchmark Schemes on MM-Fi \textbf{P3-S1} and WiPose Datasets (M: Million, G: Giga).
  }
  \label{tab:table5}
  \centering
  \normalsize
\begin{tabular}{c|ccccccl}
\hline 
 & \multicolumn{6}{c}{\textbf{MM-Fi}}\tabularnewline
\hline
 & $\textbf{PCK}_{20}$ & $\textbf{PCK}_{30}$ & $\textbf{PCK}_{40}$ & $\textbf{PCK}_{50}$$\uparrow$  & \textbf{MPJPE}$\downarrow$  & \textbf{PA-MPJPE}$\downarrow$ & \textbf{Params.}$\,$\&$\,$\textbf{FLOPs}\tabularnewline
\hline 
Wi-Pose & 48.55  & 65.06 & 75.58 & 82.41 & 158.21 & 97.72 & 5.34M$\,\,\,\,$\&$\,$ 84.31G\tabularnewline
Wi-Mose & 48.67  & 66.58  & 77.34 & 83.87 &  155.76& 95.35 & 36.20M$\,$\&$\,\,\,$245.64G\tabularnewline
WiLDAR & 44.12  & 62.58 & 72.64 & 79.26 & 170.38 & 115.64 & \textbf{1.63M}$\,\,$\&$\,$ \underline{4.91G} \tabularnewline
WiSPPN & 45.41 & 63.21 & 74.08 & 80.97 & 166.59& 110.03& 26.78M$\,$\&$\,$ 159.81G \tabularnewline
PerUnet & \underline{50.12} & \underline{67.34} & \underline{77.59} & \underline{83.56} & \underline{154.66} & \underline{98.67} & 34.51M$\,$\&$\,$ 168.52G \tabularnewline
MetaFi++ & 45.46 & 64.44 & 75.13 & 81.75 & 164.45& 106.31& 26.42M$\,$\&$\,$ 507.89G\tabularnewline
\textbf{WiLHPE} & \textbf{54.88} & \textbf{70.36} & \textbf{79.22} &\textbf{85.26} &\textbf{142.77}&\textbf{93.89}   & \underline{1.78M} \& \textbf{4.45G} \tabularnewline

\hline 
 & \multicolumn{6}{c}{\textbf{WiPose} }\tabularnewline
\hline
 & $\textbf{PCK}_{5}$ & $\textbf{PCK}_{10}$ & $\textbf{PCK}_{20}$ & $\textbf{PCK}_{50}$$\uparrow$ & \textbf{MPJPE}$\downarrow$ & \textbf{PA-MPJPE}$\downarrow$ & \textbf{Params.}$\,$\&$\,$\textbf{FLOPs}\tabularnewline
\hline 
Wi-Pose & 46.23 & 62.78 & 74.21 & 85.69 & 34.36 & 40.12  & 6.76M$\,\,\,\,$\&$\,$ 38.49G\tabularnewline
Wi-Mose & 54.65  & 66.74 & 77.12 & 88.54 & 26.48 & 31.19 & 35.75M$\,$\&$\,\,\,$116.47G \tabularnewline
WiLDAR & 36.26 & 54.38 & 72.16 & 84.32 & 55.63 &  62.60& \textbf{1.63M}$\,\,$\&$\,$ \textbf{4.90G}\tabularnewline
WiSPPN & 52.95 & 64.16 & 75.46 & 86.26 & 30.37 & 36.42  & 26.33M$\,$\&$\,$ 75.81G\tabularnewline
PerUnet & \underline{63.07} & \underline{71.77} & \underline{79.50} & \underline{88.74} & \underline{17.12} & \underline{22.64} &33.85M$\,$\&$\,$ 167.51G \tabularnewline
MetaFi++ & 53.64 & 66.72 & 76.68 & 88.62 & 28.62 & 33.72 & 25.58M$\,$\&$\,$ 502.32G\tabularnewline
\textbf{WiLHPE} & \textbf{69.79} & \textbf{82.46} & \textbf{89.41} & \textbf{94.27} & \textbf{15.85} & \textbf{19.21} & \underline{3.49M} $\,\,$\&$\,$ \underline{5.18G} \tabularnewline
\hline 
\end{tabular}
\end{table*}

\subsubsection{Results on WiPose}
We now focus on individual body parts using the $\text{PCK}_{\alpha}$ metric. As shown in Table~\ref{tab:table3b}, WiLHPE demonstrates impressive effectiveness, achieving average percentages of $94.27\%$ and $89.41\%$ for $\text{PCK}_{50}$ and $\text{PCK}_{20}$, respectively. Notably, WiLHPE exhibits exceptional results for body parts such as knees and hips (\textit{e.g} $100\%$ at $\text{PCK}_{50}$). Notably, the proposed scheme maintains commendable accuracy scores even under stricter criteria such as $\text{PCK}_{5}$ and $\text{PCK}_{10}$, with approximately $69.79\%$ and $82.46\%$, respectively. This underscores the robustness of WiLHPE's performance on specific body joints when evaluated against rigorous standards on the WiPose dataset. This accomplishment surpasses that of MM-Fi due to differences in the methods used to acquire raw signals in the two datasets. Compared to MM-Fi, the WiPose system, with its simplified configuration and lower-intensity human activities, provides higher accuracy and reliability in HPE tasks.

\subsubsection{Comparison with Benchmark Schemes} Table~\ref{tab:table5} presents results for WiLHPE and the considered benchmark schemes.
Firstly, the proposed WiLHPE scheme on the MM-Fi dataset surpasses the performance of MetaFi++, which previously achieved SOTA results in HPE tasks. WiLHPE improves $\text{PCK}_{20}$ and $\text{PCK}_{50}$ scores by approximately $4\%$ and $2\%$, respectively, and reduces MPJPE by less than $12$ mm. Notably, WiLHPE maintains significantly lower complexity, requiring fewer than $15$ times the resources compared to MetaFi++. Notably, our method on the MM-Fi dataset exhibits a complexity level similar to WiLDAR, one of the most lightweight models in human recognition tasks, with approximately $1.6$ million parameters. Nevertheless, WiLHPE offers superior performance over WiLDAR while requiring only half the number of FLOPs.

Next, on the WiPose dataset, WiLHPE outperforms other benchmark schemes in all $\text{PCK}_{\alpha}$ scores, MPJPE, and PA-MPJPE metrics. Specifically, WiLHPE shows the best performance with improvements of around $6\%$ and $16\%$ compared to PerUnet and MetaFi++ in $\text{PCK}_{5}$, respectively. Additionally, WiLHPE achieves MPJPE and PA-MPJPE results of $19.21$ mm and $15.85$ mm, respectively, significantly lower than its counterparts. Meanwhile, WiLHPE requires minimal computational resources, with only $3.49$ million parameters, slightly larger than WiLDAR, while exhibiting outstanding efficiency compared to other benchmark schemes. These comparisons underscore the remarkable effectiveness and low complexity of our approach in HPE tasks.

\subsubsection{Qualitative Result} 
Fig.~\ref{fig:figure6} shows qualitative results by providing visual representations of human skeletons captured in various scenarios from the MM-Fi dataset. Specifically, we focus on generating 2D poses and subsequently transforming them into 3D poses by adding a constant vector as the third dimension, facilitating easier visualization. The proposed WiLHPE approach consistently and reliably generates human poses across a spectrum of daily activities and rehabilitation exercises.

\subsection{Robustness Analysis}

\begin{figure}[tb]
  \centering
  \includegraphics[width=0.8\linewidth]{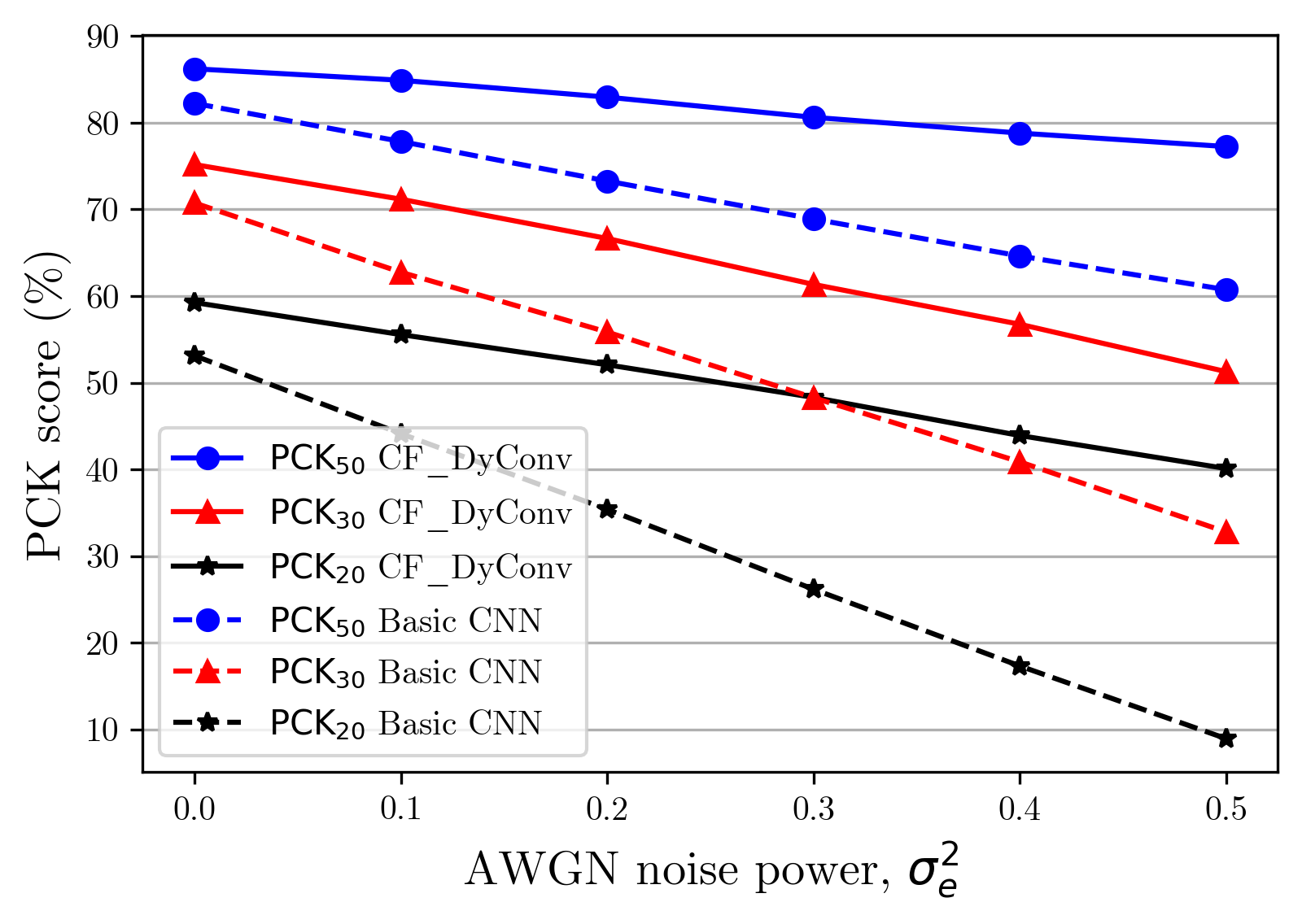}
  \caption{The PCK scores of WiLHPE versus different levels of AWGN noise.}
  \label{fig:figure7}
\end{figure}

\begin{figure}[tb]
  \centering
  \includegraphics[width=0.8\linewidth]{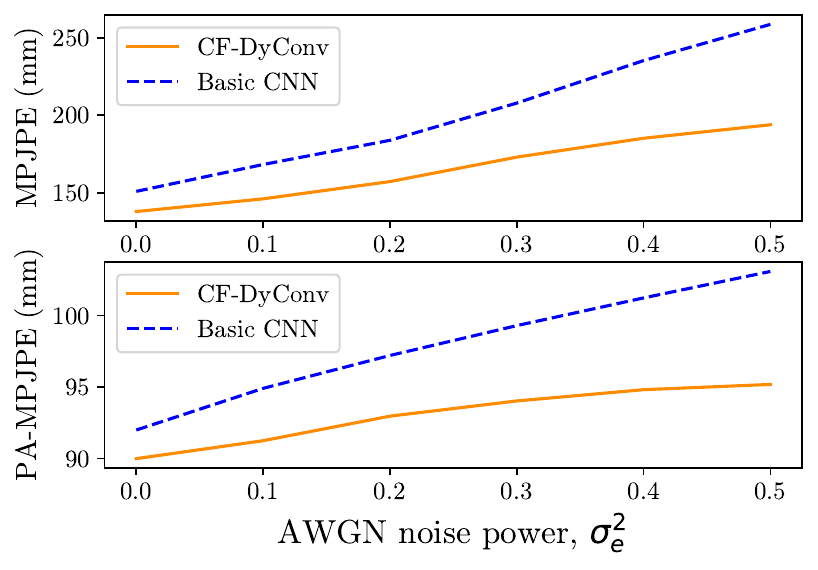}
  \caption{The MPJPE and PA-MPJPE performance versus different levels of AWGN noise.}
  \label{fig:figure8}
\end{figure}

\subsubsection{Impact of AWGN Noise} In Fig.~\ref{fig:figure7},
 we vary the values of $\sigma_{e}^2$ from $0$ to $0.5$ to assess the WiLHPE's robustness. Additionally, we implement WiLHPE with a basic CNN to highlight the adaptive capabilities of CF-DyConv in severe noise conditions.  As seen, CF-DyConv is only slightly affected by the AWGN noise and achieves good performance with accuracies of approximately $78\%$, $52\%$ and $40\%$ at $\text{PCK}_{50}$, $\text{PCK}_{30}$, and $\text{PCK}_{20}$ with $\sigma_e^2=0.5$, respectively. In contrast, the basic CNN exhibits a drastic decline as $\sigma_{e}^2$ gradually increases. In particular, the $\text{PCK}_{50}$ and $\text{PCK}_{20}$ scores of the basic CNN is around $75\%$ and $35\%$, respectively, at $\sigma_{e}^2=0.2$, dropping dramatically to approximately $60\%$ and $9\%$, respectively, as $\sigma_{e}^2$ increases to $0.5$.

Next, we evaluate the WiLHPE's robustness with the MPJPE and PA-MPJPE metrics. Fig.~\ref{fig:figure8} illustrates the response of CF-DyConv to different noise levels, revealing a marginal impact with a maximum MPJPE value below $200$ mm under the most severe condition, \textit{e.g.} ${\sigma_e^2} = 0.5$. In contrast, the basic CNN exhibits more serious susceptibility to noise, as evidenced by its MPJPE value surpassing that of our network in the worst-case scenario, reaching about $250$ mm at ${\sigma_e^2} = 0.5$. Regarding the PA-MPJPE metric, there is a notable increase of $15$ mm for the basic CNN model, rising from $90$ mm to $105$ mm as ${\sigma_e^2}$ increases from $0$ to 0$.5$. Meanwhile, the PA-MPJPE value of CF-DyConv experiences only a slight increase, reaching $95$ mm at ${\sigma_e^2} = 0.5$. These results confirm the robust adaptability of CF-DyConv in severe conditions with the presence of AWGN noise.

\subsubsection{Adversarial Attack}

\begin{figure}[tb]
  \centering
  \includegraphics[width=0.8\linewidth]{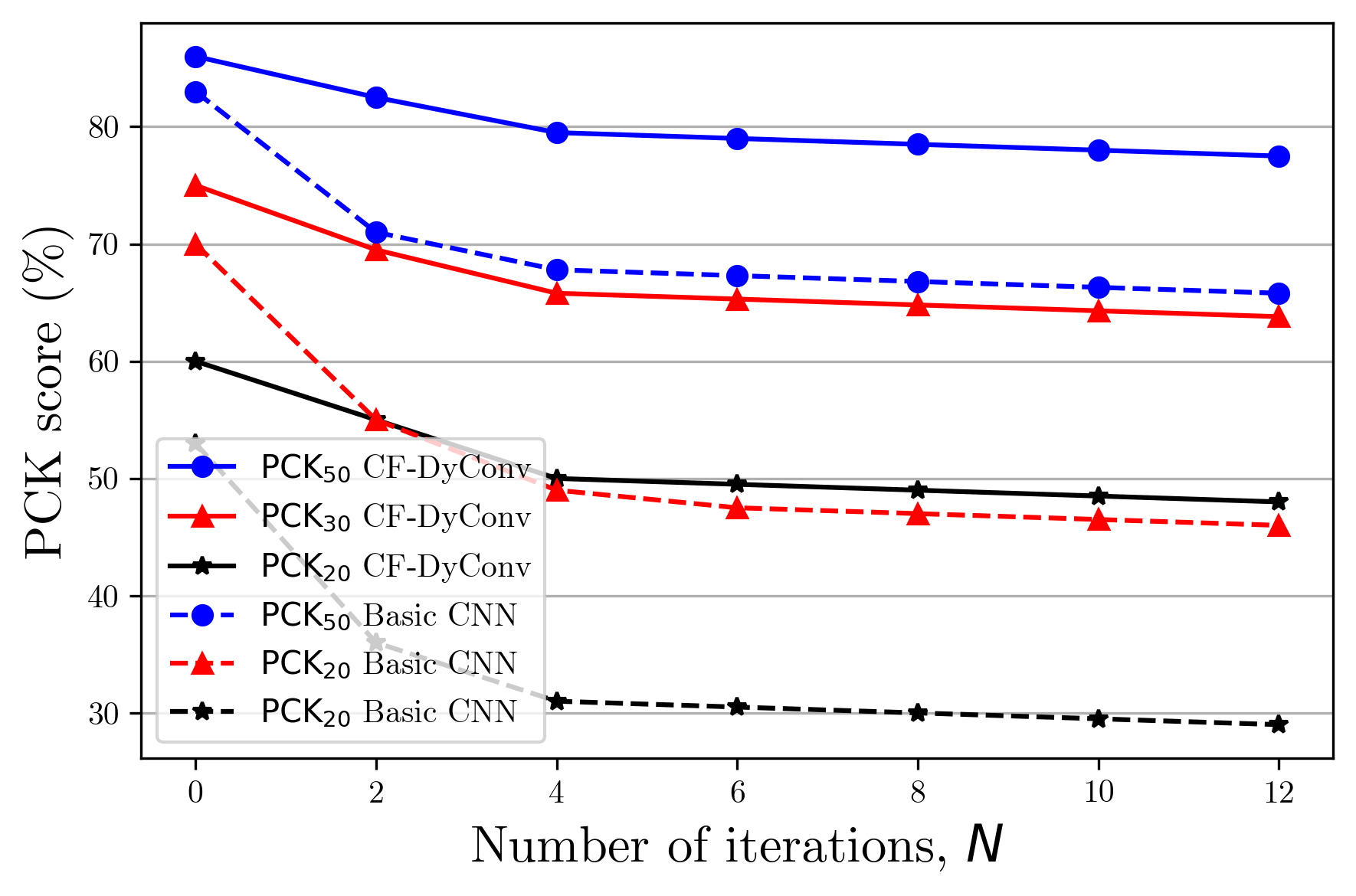}
  \caption{The PCK scores of WiLHPE under FGSM-U-N versus the number of iterations, $N$.}
  \label{fig:figure9}
\end{figure}

\begin{figure}[tb]
  \centering
  \includegraphics[width=0.8\linewidth]{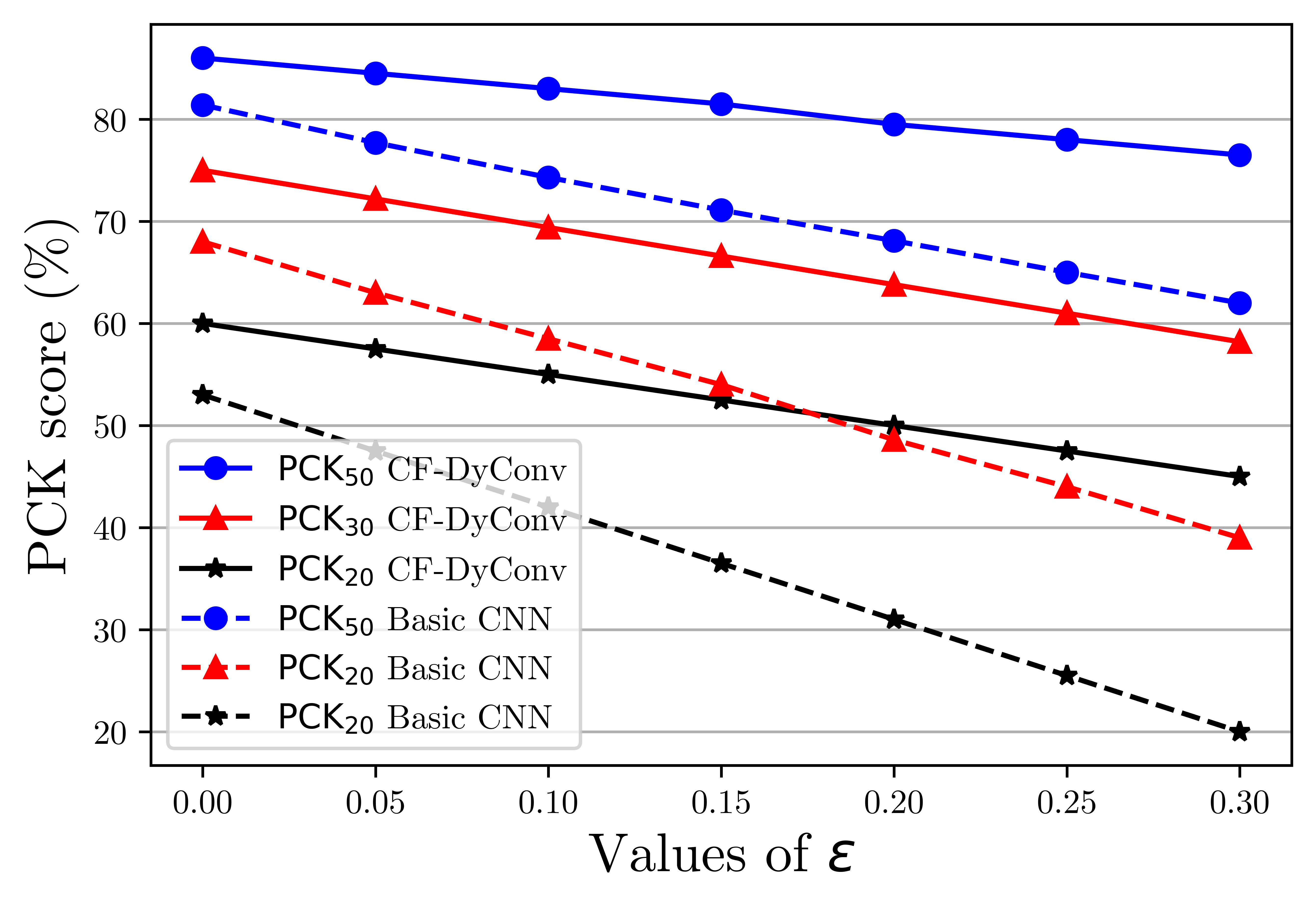}
  \caption{The PCK scores of WiLHPE under FGSM-U-N versus different values of $\varepsilon$.}
  \label{fig:figure10}
\end{figure}

To evaluate the potential impact of adversarial noise, we perform FGSM-U and FGSM-U-N on \textbf{P3-S1} MM-Fi dataset, as presented previously in Section \ref{5a}. We vary the number of iterations $N$ from $0$ to $12$. Fig. \ref{fig:figure9} shows that WiLHPE under the influence of FGSM-U-N maintains a stable $\text{PCK}_{50}$ score with a wide range of $N$ and $\varepsilon=0.2$, consistently exceeding approximately 78\% $N=10$. Furthermore, WiLHPE experiences only a slight drop in $\text{PCK}_{30}$ and $\text{PCK}_{20}$ score, decreasing by about 10\% when $N$ increases from $0$ to $12$. In contrast, adversarial noise significantly impacts the basic CNN's performance, with the $\text{PCK}{50}$ score dropping below 70\% at $N=4$. Moreover, the accuracy of this model shows a remarkable degradation in the $\text{PCK}{20}$ and $\text{PCK}{30}$ score, up to around $20\%$, as $N$ varies from $0$ to $12$. It is noteworthy that the influence of adversarial noise remains relatively consistent from $N=4$ onwards at the fixed value of $\varepsilon$.

Fig.~\ref{fig:figure10} shows that the WiLHPE's performance under the presence of adversarial noise FGSM-U-N with $N=4$ is slightly degraded as $\varepsilon$ increases. Specifically, the $\text{PCK}_{50}$ score reaches $77\%$ with $\varepsilon = 0.3$. Moreover, we observe a reduction of around $12\%$ in $\text{PCK}_{20}$ as $\varepsilon$ gradually increases from $0$ to $0.3$. Meanwhile, WiLHPE with basic CNN is seriously affected by adversarial noise FGSM-U-N. This model experiences a dramatic decline in $\text{PCK}_{50}$ as $\varepsilon$ increases (\textit{e.g} $\text{PCK}_{50}=50\%$ at $\varepsilon = 0.3$), and even worst with $\text{PCK}_{20}$. These results highlight the robust adaptability of CF-DyConv compared to basic CNN, particularly under challenging conditions such as adversarial attacks.

\subsection{Ablation Study}
To examine the impact of CF-DyConv, we use Algorithm \ref{alg:TPE} to automatically search the structural parameters of WiLHPE, resulting in optimal values for a batch size of $64$ and a convolution kernel size of $5$. 

\subsubsection{CF-DyConv}
\begin{table}
\centering
\small
\caption{Performance Comparison of CF-DyNet in WiLHPE
with Existing Selective Attention Mechanisms on the \textbf{P3-S2} MM-Fi Dataset.}
\label{tab:table7-1}

\begin{tabular}{c|ccccc}
\hline 
General Models & $\textbf{PCK}_{20}$$\uparrow$ & \textbf{MPJPE}$\downarrow$   & \textbf{Params.} & \textbf{FLOPs}\tabularnewline
\hline 
CNN+CNN &23.15& 246.59 & 1.545M & 2.256G \tabularnewline
CNN+SENet  &28.24& 237.73  & 2.138M  & 20.434G \tabularnewline
CNN+SKConv  &\underline{31.27}& \underline{210.45}  & 1.607M &  2.376G \tabularnewline
CNN+DyConv  &31.18& 214.92  & 1.759M & 4.277G \tabularnewline
\textbf{CNN+CF-DyConv}  &\textbf{43.88}& \textbf{179.24} & 1.788M  & 4.457G 
\tabularnewline
\hline 
\end{tabular}
\end{table}

We evaluate standard CNN and attention-based CNN models on the \textbf{P3-S2} MM-Fi dataset to assess cross-subject performance, where individual frequency variations influence signals even under identical postures. As shown in Table~\ref{tab:table7-1}, WiLHPE is compared with regular CNNs, SENet \cite{SENet2017}, SKConv \cite{SKNets}, and DyConv \cite{DynamicConv}. SENet derives kernel attention from global features using GAP and fully connected layers with softmax activation, whereas SKConv adapts kernel sizes but relies solely on channel-domain attention, potentially losing important information. As a result, CF-DyConv significantly outperforms SENet, improving $\text{PCK}_{20}$ by over $20\%$ and reducing MPJPE by up to $58$ mm. Compared with SKConv and DyConv, CF-DyConv further reduces MPJPE by $31$ mm and $35$ mm, respectively, demonstrating its ability to capture contextual information across both frequency and channel domains for more accurate human pose estimation.

\subsubsection{Number of Convolution Kernels}
\begin{table}
\centering
\small
\caption{WiLHPE Performance with Different Numbers of Kernels $n$ on the \textbf{P3-S1} MM-Fi Dataset.}
\label{tab:table7-2}
\begin{tabular}{c|ccccc}
\hline 
\textbf{No. kernels $n$} & $\textbf{PCK}_{20}$$\uparrow$ & \textbf{MPJPE}$\downarrow$   & \textbf{Params.} & \textbf{FLOPs} \tabularnewline
\hline 
1 &53.46& 144.65 & 1.62M & 4.26G \tabularnewline
2  &\underline{54.91}& \underline{143.24} &  1.78M  & 4.45G \tabularnewline
3  &\textbf{54.95} & \textbf{142.77}  & 1.86M &  4.67G \tabularnewline
4  &54.93& 143.84  & 1.98M & 4.77G \tabularnewline
\hline 
\end{tabular}
\end{table}

From \eqref{eq:post-process4}, we now evaluate the impact of the number of convolution kernels $n$. For simplicity, we select $r = 32$ and $\tau = 34$. We consider different values of $n$, ranging from 1 to 4. As shown in Table \ref{tab:table7-2}, WiLHPE achieves optimal results in terms of $\text{PCK}_{20}$ with $n=3$ and $n=4$, surpassing $54.95\%$. This implies that WiLHPE's performance remains consistent with $n \geq 3$. However, due to increased model complexity with higher $n$ values, we select $n=3$ in our simulations.

\subsubsection{Reduction Ratio and Temperature}
\begin{table}[]
    \caption{Sensitivity Analysis of Ratio $(r)$ and Temperature $(\tau)$
on the \textbf{P3-S1} MM-Fi Dataset.}
    \centering
    \footnotesize
    \begin{tabular}{cccccc}
\hline
$\textbf{Ratio}$ $r$ & $\textbf{Temp. }\textbf{\ensuremath{\tau}}$ & $\textbf{PCK}_{50}$ & $\textbf{PCK}_{20}\uparrow$ & $\textbf{MPJPE\ensuremath{\downarrow}}$ & $\textbf{Params.}$\tabularnewline
\hline 
\multirow{4}{*}{4} & 1 & 85.292  & 54.348  & 146.539 & \multirow{4}{*}{1.35M}\tabularnewline
 & 10 & 85.553  & 54.166  & 146.281 & \tabularnewline
 & 20 & 85.246  & 54.802  & 145.825 & \tabularnewline
 & 30 &  85.294& 54.781  & 145.585 & \tabularnewline
\hline 
\multirow{4}{*}{8} & 1 & 84.899  & 53.127  &145.562  & \multirow{4}{*}{1.56M}\tabularnewline
 & 10 & 84.792 & 53.904  & 145.385 & \tabularnewline
 & 20 & 84.870 & 53.426 &  145.196& \tabularnewline
 & 30 & 84.669 &  53.528&  144.680& \tabularnewline
\hline 
\multirow{4}{*}{\textbf{16}} & 1 & 85.293  &  54.128  & 144.126 & \multirow{4}{*}{1.78M}\tabularnewline
 & 10 & 85.338 & 54.115 & 143.706 & \tabularnewline
 & 20 & 85.512 & 54.385 & 143.124 & \tabularnewline
 & \textbf{30} & \textbf{85.962} &  \textbf{54.882}& \textbf{142.774} & \tabularnewline
\hline 
\multirow{4}{*}{32} & 1 & 84.585  & 53.041 & 145.839 & \multirow{4}{*}{1.92M}\tabularnewline
 & 10 & 84.942 & 53.636 &  145.428 & \tabularnewline
 & 20 & 84.877 & 53.456 & 145.271 & \tabularnewline
 & 30 & 85.232 &  53.982 & 144.814 & \tabularnewline
\hline 
\end{tabular}
      \label{tab:table7-3}
\end{table}

Finally, the impact of the reduction ratio $r$ and temperature $\tau$ is given in Table \ref{tab:table7-3}. During the excitation stage of our attention module, the reduction ratio $r$ determines the size of the convolutional layer and adjusts the complexity of WiLHPE. We conduct experiments on the MM-Fi dataset with $r=\{4, 8, 16, 32\}$ to identify the optimal ratio.
Results from Table \ref{tab:table7-3} show that increasing the reduction ratio $r$ reduces the model's complexity without compromising performance. The optimal balance between performance and complexity is achieved with $r=16$. This implies that adjusting the reduction ratio effectively reduces model complexity and enhances computational efficiency while retaining crucial information. Moreover, this method of dimensionality reduction simplifies the training process by decreasing the number of parameters to optimize, thereby mitigating issues related to vanishing or exploding gradients.

Table \ref{tab:table7-3} also shows the impact of the temperature value $\tau$ on kernel attention, which regulates the sparsity of the attention mechanism. We experiment with temperature values of $\tau = \{1, 10, 20,  30\}$ to determine the optimal temperature for the proposed network. As can be seen, a temperature value of $\tau = 30$ achieves a well-balanced network with high performance and low complexity. Consequently, we adopt $\tau = 30$ as the default temperature value to present the results of WiLHPE.

\section{Conclusions\label{subsec:sec6}}
In this paper, we have proposed WiLHPE, a novel WiFi-based model designed to derive human pose landmarks by analyzing raw WiFi signals. WiLHPE achieved state-of-the-art accuracy while maintaining a lightweight model structure, with significantly fewer parameters compared to existing models in performing HPE tasks. The proposed model integrates a kernel attention mechanism that dynamically adjusts the kernel size based on input data characteristics. Furthermore, we have implemented the TPE algorithm to optimize critical hyperparameters, enabling WiLHPE to achieve optimal performance without requiring extensive hyperparameter searches. Rigorous validation on the MM-Fi and WiPose datasets demonstrated WiLHPE's resilience and adaptability across various experimental settings. Notably, our model has shown significant performance improvements even under severe conditions. These achievements have paved the way for real-world deployment of WiLHPE, aimed at enhancing daily experiences for individuals.

\begingroup
\balance
\bibliography{refs}
\bibliographystyle{IEEEtran}
\endgroup

\end{document}